\documentclass[runningheads]{llncs}
\usepackage{svg}
\usepackage{tikz}
\usepackage{xcolor}
\usepackage{multirow}
\usepackage{pifont}
\newcommand{\cmark}{\ding{51}} 
\newcommand{\xmark}{\ding{55}} 
\usepackage[T1]{fontenc}
\usepackage{graphicx,verbatim}
\usepackage{hyperref}
\usepackage{color}

\usepackage{amssymb}

\begin{document}
\title{PRECISE-AS: Personalized Reinforcement Learning for Efficient Point-of-Care Echocardiography in Aortic Stenosis Diagnosis}

\author{Armin Saadat\inst{1} \and
Nima Hashemi\inst{1} \and
Hooman Vaseli\inst{1} \and
Michael Y. Tsang\inst{3} \and
Christina Luong\inst{3} \and
Michiel Van de Panne\inst{2} \and
Teresa S. M. Tsang\inst{3} \and
Purang Abolmaesumi\inst{1}} 


\authorrunning{A. Saadat et al..}

\institute{Department of Electrical and Computer Engineering, The University of British Columbia, Vancouver, BC, Canada \\
\email{\{arminsdt,purang\}@ece.ubc.ca}\\ \and
Department of Computer Science, The University of British Columbia, Vancouver, BC, Canada \and
Vancouver General Hospital, Vancouver, BC, Canada \footnote{T. S.M. Tsang and P. Abolmaesumi are joint senior authors.}}

\maketitle              

\begin{abstract}
Aortic stenosis (AS) is a life-threatening condition caused by a narrowing of the aortic valve, leading to impaired blood flow. Despite its high prevalence, access to echocardiography (echo)—the gold-standard diagnostic tool—is often limited due to resource constraints, particularly in rural and underserved areas. Point-of-care ultrasound (POCUS) offers a more accessible alternative but is restricted by operator expertise and the challenge of selecting the most relevant imaging views. To address this, we propose a reinforcement learning (RL)-driven active video acquisition framework that dynamically selects each patient's most informative echo videos. Unlike traditional methods that rely on a fixed set of videos, our approach continuously evaluates whether additional imaging is needed, optimizing both accuracy and efficiency. Tested on data from 2,572 patients, our method achieves 80.6\% classification accuracy while using only 47\% of the echo videos compared to a full acquisition. These results demonstrate the potential of active feature acquisition to enhance AS diagnosis, making echocardiographic assessments more efficient, scalable, and personalized. Our source code is available at: \url{https://github.com/Armin-Saadat/PRECISE-AS}.


\end{abstract}

\section{Introduction}
Aortic stenosis (AS)~\cite{as_intro} is a life-threatening heart valve disease marked by progressive leaflet thickening and calcification, restricting blood flow from the left ventricle. If untreated, severe AS has a five-year mortality rate of 67\%~\cite{strange20191851}. Early detection is critical but remains inaccessible due to screening limitations.

Echocardiography (echo) is the gold-standard for AS assessment, but its use is constrained by a shortage of trained personnel, leading to long wait times~\cite{munt2006access,sanfilippo2005guidelines}. Point-of-care ultrasound (POCUS) offers a more accessible alternative by enabling non-specialists to perform focused cardiac imaging at the bedside~\cite{luong2018focused}. However, POCUS exams lack comprehensive Doppler imaging and depend on operator expertise. Machine learning (ML) can bridge this gap by automating data acquisition and interpretation, assisting operators in capturing optimal echo views and determining the need for additional imaging-enhancing accuracy while reducing acquisition time.

Active feature acquisition (AFA) has been explored to optimize test-time data collection, with methods such as $L_1$ regularization enforcing sparsity at the population level and patient-specific techniques utilizing decision trees, mutual information-based selection, or Markov Decision Processes (MDP). However, AFA has been largely limited to low-dimensional data, with high-dimensional medical imaging applications restricted to pixel-level sampling in static images~\cite{akrout2018improving,akrout2019,gabriel2022rl,yuxin2015seqinfo}, leaving video-based dynamic acquisition unexplored.

We introduce PRECISE-AS, the first active video acquisition framework for echocardiography, leveraging reinforcement learning (RL) to dynamically select the most informative echo videos for each patient. Our method sequentially acquires data until diagnostic confidence is achieved, optimizing the trade-off between accuracy and acquisition time. Unlike static approaches that rely on predefined echo views, PRECISE-AS constructs personalized diagnostic pathways, ensuring patient-specific optimization of video acquisition. 

Tested on a dataset of 2,572 patients, PRECISE-AS achieves state-of-the-art accuracy in AS detection while significantly reducing the number of required echo videos. These results demonstrate the potential of AFA in high-dimensional medical imaging, enabling cost-effective, scalable, and patient-tailored echocardiographic assessments at the point of care.

\section{Background}
\subsubsection{Active Feature Acquisition.}
AFA frames the data acquisition problem as a discrete-time decision-making process~\cite{yin2020reinforcement} and operates under the premise that not all features are always available. The process begins with an empty feature set, sequentially acquiring additional features until it reaches a target prediction accuracy or the acquisition budget is exhausted. AFA seeks to balance prediction error against acquisition cost~\cite{li2021towards}. Approaches to AFA include greedy algorithms, which iteratively select the feature with the highest marginal information gain~\cite{akrout2018improving,covert2023learning,zannone2019odin}, and MDP-based methods that employ reinforcement learning to optimize the entire sequence of feature acquisitions rather than just the next immediate step~\cite{gabriel2022rl,muyama2024deep,yu2023deep}.

\subsubsection{Markov Decision Process.}
An MDP~\cite{Littman2001} is defined by  $(S, S_{end}, A, R, T)$, where $S$ is the set of states, $S_{end} \subset S$ the terminal states, and $A$ the set of actions. The transition function $T: S \times A \times S \rightarrow \mathbb{R}$ gives the probability of moving to a new state given the current state and action, while the reward function $R: S \times A \times S \rightarrow \mathbb{R}$ assigns rewards based on these transitions. An episode is a sequence of states, actions, and rewards from an initial state to a terminal state. A policy $\pi: S \rightarrow A$ maps states to actions, and the goal is to find the optimal policy $\pi^*$ that maximizes the expected cumulative reward.

\section{Method}
\begin{figure}[t]
\centering
\includegraphics[width=\textwidth]{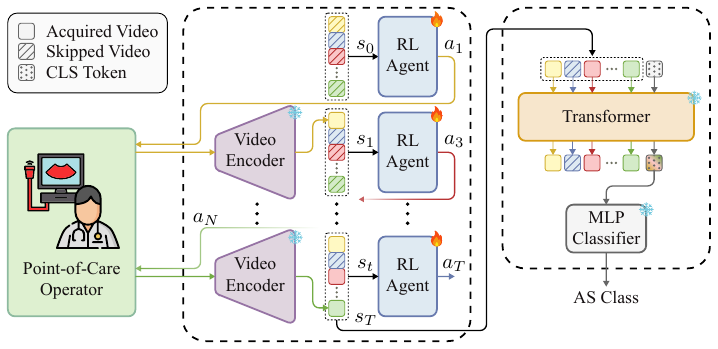}
\caption{Overview of the structure. Each action $a_i$ acquires the $i$-th video and updates the state. At termination ($a_T$), the final state $s_T$ is passed on for classification.}
\label{fig:architecture}
\end{figure}

PRECISE-AS (Fig.~\ref{fig:architecture}) consists of three key components. 1) A video encoder $f(.)$ that maps an echo video $x_i \in \mathbb{R}^{H \times W \times 3}$ to feature $f(x_i) \in \mathbb{R}^D$, where $H$, $W$ correspond to the height and width of the frames and $D$ is the embedding size of the features. 2) An RL agent that iteratively selects encoded videos for each patient until it terminates the acquisition process. 3) A classifier $g(.)$ that integrates the selected encoded videos and makes a diagnosis. The encoder and classifier are trained and frozen to be used for training the RL agent.
The AS severity label is predicted as follows:
\begin{equation}
\label{eq:classification}
y^{\mathrm{pred}} \;=\; \arg\max \bigl(g(z_1, z_2, \dots, z_N)\bigr), 
\quad\mathrm{where}\quad
z_i \;=\; \bigl[m_i \cdot f(x_i)\bigr].
\end{equation}
Here, $N$ is the maximum number of videos per study, $m_i$ is a boolean scalar indicating whether $x_i$ is selected, and $\cdot$ denotes a dot product. The following sections describe each component in detail.

\subsection{Feature Extraction}
PRECISE-AS utilizes the open-source ProtoASNet~\cite{vaseli2023protoasnet} as its feature extractor model. ProtoASNet, currently the state-of-the-art for AS classification from B-mode echo videos, is a prototypical network that classifies AS severity while remaining transparent in its decision-making. It learns spatio-temporal prototypes—representative clips highlighting calcification and restricted aortic valve leaflet motion—and compares new inputs against these reference points. By precisely locating clinically relevant features in each video frame, ProtoASNet offers an interpretable approach in which predictions hinge on a handful of evidence-based prototypes rather than opaque, large-scale parameters.

Class-wise similarity scores are computed by gauging how closely an input embedding aligns with each prototype, and a fully connected layer then weights these similarity scores to yield a probability for each AS severity category. This design ensures that the “why” behind each classification remains grounded in clinically significant patterns. ProtoASNet also incorporates aleatoric uncertainty by introducing prototypes that capture ambiguous or poor-visibility regions, flagging uncertain cases when necessary. Rather than using its final classification output, we specifically take the vectors feeding into the final fully connected layer as our feature representation for subsequent steps in our analysis. We train this feature extractor independently before the rest of the framework, then freeze its learned parameters to produce a fixed feature set for training the classifier.

\subsection{Feature Selection} 
PRECISE-AS formulates feature selection as an MDP. Each state $s \in \mathbb{R}^{N \times D}$ represents the current subset of selected videos. In this state space, chosen videos retain their feature embeddings, while masked vectors replace unselected videos. The action space consists of $\{a_T, a_1, \ldots, a_N\}$, where $a_i$ selects video $x_i$ and $a_T$ terminates the acquisition process. The start state ($s_0$) is zero-initialized. As shown in Eq.~(\ref{eq:transition}), 
taking action $a_i$ updates the $i$-th entry of the state to $f(x_i)$:
\begin{equation}
\label{eq:transition}
s_{t+1} = s_{t}.\mathrm{copy()}, 
\quad
s_{t+1}[i] = f(x_i).
\end{equation}
Similar to~\cite{gabriel2022rl}, a sparse reward scheme is adopted. Non-terminal states receive a reward of 0, while terminal states receive the following reward:
\begin{equation}
\label{eq:reward}
R(s_T) = \mathbf{1}\bigl(y_{s_T}^{\mathrm{pred}} = y^{\mathrm{true}}\bigr) \;-\; \lambda \sum_{i \in s_T} c_i,
\end{equation}
where \(s_T\) denotes a terminal state, and \(\mathbf{1}(\cdot)\) is the indicator function that returns 1 if \(y_{s_T}^{\mathrm{pred}} = y^{\mathrm{true}}\) (accuracy) or 0 otherwise. The term \(c_i\) is the cost of acquiring the \(i\)-th video in state \(s\), and \(\lambda\) balances acquisition cost against classification accuracy. To obtain \(y_{s_T}^{\mathrm{pred}}\), we feed the terminal state into a frozen pre-trained classifier. The agent can take at most \(N+1\) actions, up to \(N\) for acquiring videos and one final termination action \(a_T\). If the agent uses all \(N+1\) actions without invoking \(a_T\), the episode automatically ends with zero reward.

PRECISE-AS uses Q-learning~\cite{watkins1992q} to get the optimal policy and maximize the expected reward. Given the continuous, high-dimensional nature of our state space, we employ Double Deep Q-Networks (DDQN)~\cite{van2016deep}, an enhancement of standard DQN designed to mitigate overestimation bias and stabilize training.

\subsection{Feature Integration and Classification}
The AS severity classifier has two objectives: (1) providing the reward signal during RL agent training, and (2) making predictions using the subset of videos selected by the agent at inference time. Consequently, the classifier must adapt to varying numbers of available videos and remain robust when some views are missing, while leveraging study-level aggregation to enhance performance beyond individual video-level predictions.

To accomplish this, we frame the problem similarly to natural language processing. Each echo video is treated as a “token,” and we maintain a maximum of \(N\) tokens to match the maximum number of echo videos per study. Missing tokens correspond to unselected videos. To ensure positional awareness, we use a consistent ordering of echo views and apply positional encodings, so the model understands which token corresponds to which view. This setup naturally aligns with Transformer~\cite{vaswani2017attention} encoders, where each sequence of tokens (videos) is passed through the encoder along with a classification token (CLS) that aggregates information at the study level.

During training, we improve the classifier's robustness to missing data by randomly masking 50\% of the video tokens in each epoch. For any masked token (i.e., when $m_i = 0$), its features $f(x_i)$ are replaced with zeros, as shown in Eq.~(\ref{eq:classification}). We also apply attention masking to ensure that the CLS token ignores these masked tokens, compelling the model to rely on the remaining unmasked videos for classification. As a result, the classifier becomes flexible, delivering reliable predictions even with incomplete data. After training, the classifier’s parameters are fixed to be used in training the RL agent.

At inference time, the classifier processes a terminal state $s_T \in \mathbb{R}^{N \times D}$, interpreting it as a sequence of $N$ tokens. Any videos the RL agent did not select are masked out (i.e., zeroed), and attention masking ensures that only the acquired videos contribute to the final classification. This design allows the system to produce accurate and efficient AS severity assessments tailored to each patient’s available echo data.

\section{Experiments and Results}
\subsection{Datasets}
We conducted our experiments on a private AS dataset sourced from an echo study database at a tertiary care hospital, following institutional review board approval. Echo videos were acquired using Philips iE33, Vivid i, and Vivid E9 systems. AS severity was classified by a Level III echocardiographer using standard Doppler guidelines [8], and only cases with consistent Doppler measurements were retained. A view-detection algorithm~\cite{liao2019modelling} automatically identified parasternal long-axis (PLAX) and short-axis (PSAX) cine clips, which were then re-screened by a Level III echocardiographer to remove misclassifications. 
The final dataset comprises 2,572 patient studies, including 5,055 PLAX and 4,062 PSAX cine clips. The data were split into training, validation, and test sets (80-10-10), ensuring patient-level exclusivity. Each study was labeled with Doppler-derived severity levels, resulting in 1088 normal, 575 early, and 909 significant AS cases. For consistency, we randomly selected two PLAX and two PSAX videos per patient, resulting in four clips per study, ordered as [PLAX$_1$, PLAX$_2$, PSAX$_1$, PSAX$_2$]. 
This fixed ordering gives each echo view a consistent meaning.
Each video is assigned a unit cost, as all clips are captured using the same imaging modality, making the total cost directly proportional to the number of videos acquired.

\subsection{Implementation Details}
For feature extraction, we train ProtoASNet according to the implementation details outlined in the original paper. We implement DDQN using two three-layer, fully connected networks for feature selection. The RL agent is trained for 50 epochs using MSE loss. We apply a discount factor of $1.0$ to treat immediate and future rewards equally, as each additional echo video inherently carries its own cost. The best model is selected based on its validation accuracy. We utilize a Transformer with six encoder layers, eight attention heads, and a 256-dimensional feed-forward layer for classification. The network is trained using cross-entropy loss for 50 epochs. The framework is implemented in PyTorch, and models are trained on a single 16 GB NVIDIA Tesla V100 GPU. 

\subsection{Evaluations}
\subsubsection{Quantitative Assessment.}

\begin{table}[t]
\caption{
Quantitative metrics include balanced accuracy (bACC), weighted F1, and balanced mean absolute error (bMAE). bMAE is computed as the average MAE over classes (labels 0, 1, 2 for no, early, and significant AS). Study-level results are obtained by averaging video prediction probabilities per study, except for PRECISE-AS, which inherently integrates study-level information. Best results are in bold. 
}
\label{tab:results}
\centering
\begin{tabular}{c|ccc|cc}
\multirow{2}{*}{Method}                      
& \multicolumn{3}{c}{Study-level (N=252)}
\vline
& \multirow{2}{*}{Video-based}
& \multirow{2}{*}{\shortstack{Acquired \\ Videos}}
\\ 
& \multicolumn{1}{c}{~~bACC$\uparrow$~~} 
& \multicolumn{1}{c}{~~~~~F1$\uparrow$~~~~~} 
& \multicolumn{1}{c}{~~bMAE$\downarrow$~~}
\vline
\\ 
\hline 
\hline
ProtoPNet~\cite{chen2019ProtoPNet} 
& 70.9(4.7)  
& 0.69(.07)
& 0.32(.05)   
& \xmark
& 100\%  
\\ 
XProtoNet~\cite{XprotoNet}
& 73.8(0.8)    
& 0.74(.01)   
& 0.29(.01) 
& \xmark
& 100\%  
\\
ProtoASNet (Image)~\cite{vaseli2023protoasnet}
& 73.9(3.5)  
& 0.74(.04)
& 0.29(.04) 
& \xmark
& 100\%  
\\ 
Huang et al.~\cite{huangTMED2Dataset2022} 
& 74.7(1.6)
& 0.75(.02)
& 0.28(.02)  
& \xmark 
& 100\%   
\\
\hline
Ahmadi et al.~\cite{ahmadi2023transformer}
& 76.9(1.7)   
& 0.77(.02)  
& 0.25(.02) 
& \cmark
& 100\%  
\\
XProtoNet (Video)~\cite{XprotoNet}
& 77.2(1.4)   
& 0.77(.01)  
& 0.25(.02) 
& \cmark
& 100\%  
\\
Ginsberg et al.~\cite{AS_Tom}
& 78.3(1.6)   
& 0.78(.01)
& 0.24(.02) 
& \cmark
& 100\%  
\\
ProtoASNet~\cite{vaseli2023protoasnet}
& 80.0(1.1)  
& 0.80(.01)  
& 0.22(.01)
& \cmark
& 100\%  
\\
ProtoASNet+RT4U~\cite{vaseli2023protoasnet}
& 80.1(1.4)  
& 0.80(.01)  
& 0.22(.01)
& \cmark
& 100\%  
\\ \hline
PRECISE-AS (w/o RL)
& \textbf{80.6(0.5)}
& \textbf{0.83(.01)}
& \textbf{0.20(.01)}
& \cmark
& 100\%
\\
PRECISE-AS (with RL)
& \textbf{80.6(0.8)}
& \textbf{0.83(.01)}
& \textbf{0.20(.01)}
& \cmark
& \textbf{47\%} 
\end{tabular}
\end{table}

\begin{figure}
\includegraphics[width=\textwidth]{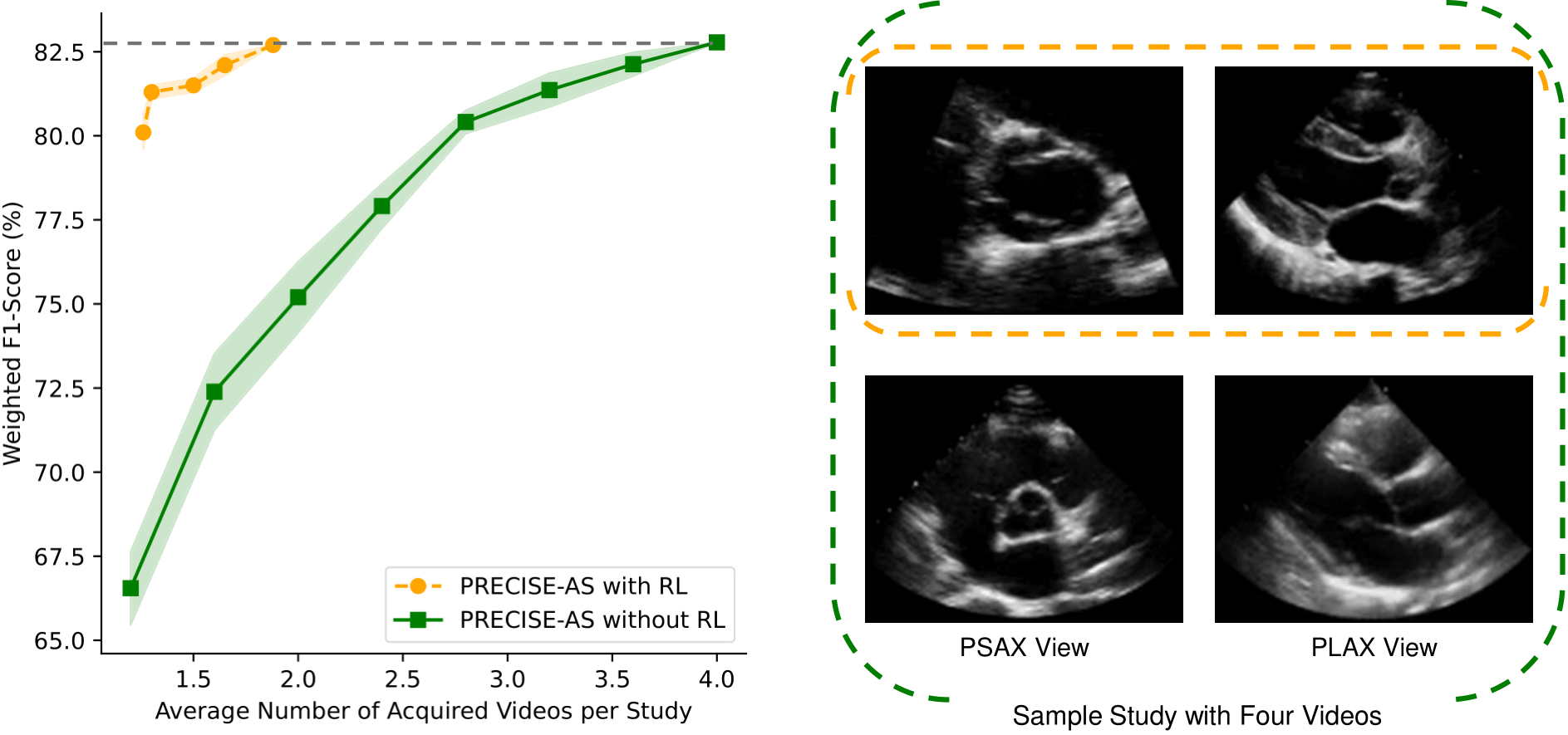}
\caption{(Left) F1 score versus the average number of acquired videos per study. Each yellow data point corresponds to a specific cost-coefficient. The solid lines represent the mean performance, while the shaded regions indicate one standard deviation. (Right) A sample study consisting of four echo videos. PRECISE-AS only requires the top two videos to achieve the same patient-level accuracy as the whole study.} 
\label{fig:acc}
\end{figure}

Table~\ref{tab:results} compares the performance of PRECISE-AS for classifying AS severity against both image-based and video-based baselines.
Video-based models outperform the image-based ones because they capture the dynamic motion of the aortic valve—a key indicator of AS severity. PRECISE-AS fully exploits this video-level information by employing the video-based ProtoASNet as its encoder.
For the baselines, study-level predictions are obtained by averaging the class probabilities predicted at the image or video level. In contrast, PRECISE-AS intrinsically operates at the study level by using an attention mechanism to enrich a CLS token that encapsulates study-level information.
This dual advantage explains why PRECISE-AS (without RL) outperforms the baselines when acquiring full studies. Moreover, incorporating RL to actively and efficiently select videos maintains performance while reducing the average number of acquired videos per study to 47\% of a full acquisition—as indicated by PRECISE-AS (with RL).
Fig.~\ref{fig:acc} provides a detailed comparison between with and without RL, showing that using RL for video selection as inputs to the classifier, consistently has a better performance-to-number-of-videos ratio, and ultimately obtains the best performance using fewer videos. This means that to get the best performance, PRECISE-AS w/o RL must select 4 videos per study on average, while PRECISE-AS with RL only needs around 2 videos, as shown in a sample study in Fig.~\ref{fig:acc}. Although PRECISE-AS fails to show statistically significant improvement over ProtoASNet (p > 0.05), it matches in evaluation metrics while using less than 50\% of the echo videos.

\subsubsection{Qualitative Assessment.}

\begin{figure}
\includegraphics[width=\textwidth]{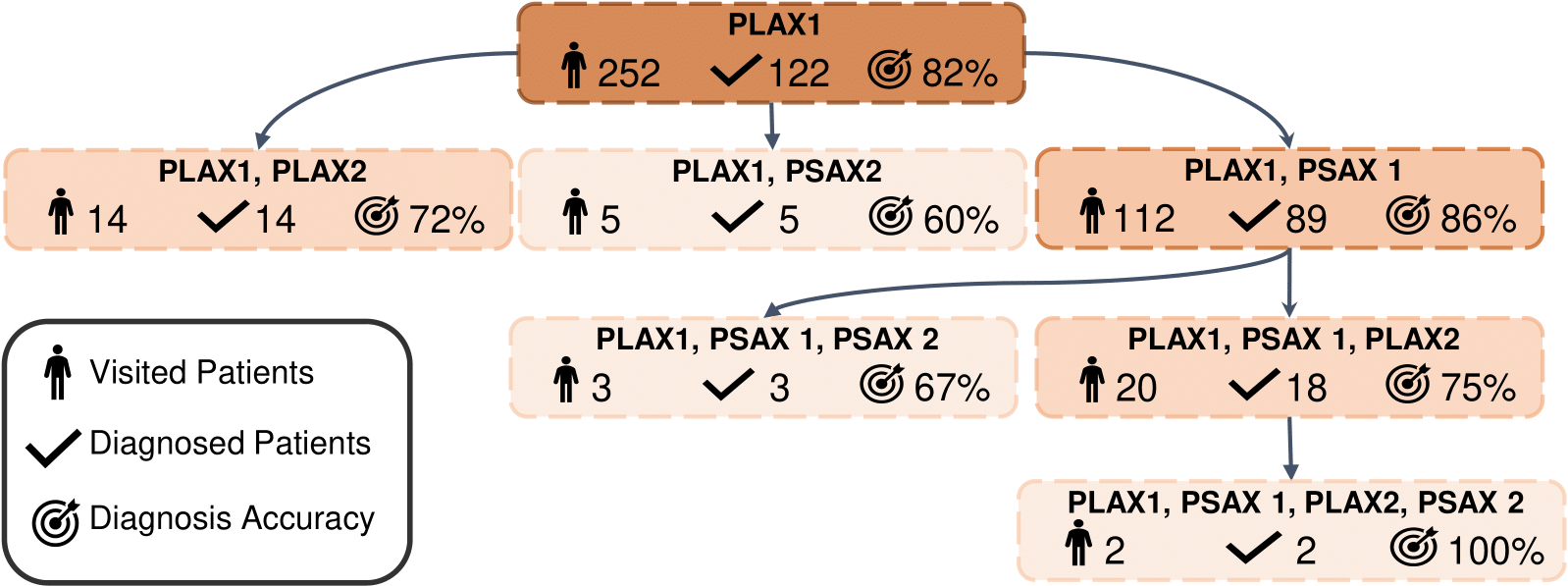}
\caption{Personalized diagnostic pathways for AS classification. Each node represents a state defined by the set of acquired videos and shows how many patients reached it. Directed edges illustrate the sequential video acquisitions leading to a final diagnosis.}
\label{fig:pathway}
\end{figure}

Active data acquisition offers critical insights into the decision pathways leading to a final diagnosis. These insights can inform AS diagnostic guidelines in healthcare by emphasizing the statistical significance of key features and modalities. As illustrated in Fig.~\ref{fig:pathway}, the optimal starting point is to acquire a PLAX view, followed by a PSAX video, thereby enabling the classifier to examine the aortic valve from two complementary perspectives. We hypothesize that when two videos from the same view are selected, the first is suboptimal.
Moreover, the agent adheres to implicit data acquisition rules as it terminates the process and avoids repetitive actions within an episode.

\subsubsection{Ablation Study.}
\begin{table}[t]
\caption{
Ablation study of the acquisition cost coefficient ($\lambda$) demonstrating the trade-off between acquisition efficiency and classification performance.}
\label{tab:ablation}
\centering
\begin{tabular}{c|ccc|cc}
\multirow{2}{*}{\shortstack{Cost \\ Coefficient}}
& \multicolumn{3}{c}{Study-level (N=252)}\vline
& \multicolumn{2}{c}{Acquired-Videos }
\\ 
& \multicolumn{1}{c}{~~bACC$\uparrow$~~} 
& \multicolumn{1}{c}{~~~~~F1$\uparrow$~~~~~} 
& \multicolumn{1}{c}{~~bMAE$\downarrow$~~}\vline
& \multicolumn{1}{c}{~~Ratio$\downarrow$} 
& \multicolumn{1}{c}{~~Count$\downarrow$}
\\ 
\hline 
\hline
w/o RL
& 80.6(0.5)
& 0.83(.01)
& 0.20(.01)
& 100\%  
& 4.00
\\
0.001
& \textbf{80.6(0.8)}
& \textbf{0.83(.01)}
& \textbf{0.20(.01)}
& 47\%  
& 1.88
\\
0.01
& 79.9(0.8)
& 0.82(.01)
& 0.21(.01)
& 41\% 
& 1.64
\\
0.1
& 79.0(0.6)
& 0.81(.01)
& 0.21(.01)
& 37\%  
& 1.50
\\
0.2
& 78.5(0.7)
& 0.81(.01)
& 0.21(.01)
& 32\% 
& 1.30
\\
0.25
& 77.1(0.6)
& 0.80(.01)
& 0.23(.02)
& 31\%
& 1.26
\end{tabular}
\end{table}

The cost coefficient ($\lambda$) in Eq.~\ref{eq:reward} plays a crucial role in training the RL agent. Increasing $\lambda$ causes the model to select fewer videos, which in turn reduces performance (see Table~\ref{tab:ablation}). If $\lambda$ is set excessively high, the RL agent terminates the acquisition process immediately without selecting any videos.

\section{Conclusion}
PRECISE-AS represents a significant advancement in automated, patient-specific echocardiographic assessment, addressing critical barriers to efficient and accessible AS diagnosis. By leveraging reinforcement learning for active video acquisition, our approach reduces unnecessary imaging while maintaining state-of-the-art diagnostic accuracy. This personalized, adaptive strategy streamlines POCUS workflows, enabling faster, more targeted AS evaluations—a crucial step toward improving early detection and timely intervention, particularly in resource-limited settings. By demonstrating the feasibility of active feature acquisition in high-dimensional medical imaging, PRECISE-AS lays the groundwork for scalable, intelligent echocardiography, ultimately enhancing clinical decision-making, operational efficiency, and patient outcomes.

\begin{credits}
\subsubsection{\ackname} This research was supported in part by the Canadian Institutes of Health Research (CIHR) and the Natural Sciences and Engineering Research Council of Canada (NSERC), and through computational resources and services provided by Advanced Research Computing at the University of British Columbia.

\subsubsection{\discintname}
The authors have no competing interests to declare.
\end{credits}

\bibliographystyle{splncs04}
\bibliography{paper-5008}

\end{document}